\documentclass[10pt,twocolumn,letterpaper]{article}

\usepackage[pagenumbers]{cvpr} 

\usepackage{graphicx}
\usepackage[skip=0pt]{caption}
\usepackage{amsmath}
\usepackage{amssymb}
\usepackage{booktabs}
\usepackage{balance}
\usepackage{enumitem}

\usepackage{graphics}
\usepackage{xspace}
\usepackage{color}
\usepackage{bm}

\usepackage{siunitx}
\usepackage{balance}
\usepackage{lmodern}
\usepackage{tabularx}
\usepackage{booktabs}
\usepackage{makecell}
\usepackage[hyphens]{url}
\usepackage{amssymb}

\usepackage[pagebackref,breaklinks,colorlinks]{hyperref}

\usepackage[capitalize]{cleveref}
\crefname{section}{Sec.}{Secs.}
\Crefname{section}{Section}{Sections}
\Crefname{table}{Table}{Tables}
\crefname{table}{Tab.}{Tabs.}

\usepackage{wrapfig}

\newcolumntype{P}[1]{>{\raggedright\arraybackslash}m{#1}}%
\newcolumntype{C}[1]{>{\centering\arraybackslash}m{#1}}%
\newcolumntype{R}[1]{>{\raggedleft\arraybackslash}m{#1}}%

\usepackage{xcolor}

\usepackage[colorlinks]{hyperref} 

\definecolor{guohaocolor}{rgb}{0, 0.8196, 0}

\definecolor{xucongcolor}{rgb}{0.73725, 0.6588, 0.0705} 

\usepackage{textcomp}
\usepackage{gensymb}
\usepackage{diagbox}

\usepackage[defaultcolor=red]{changes}   

\usepackage{etoolbox}
\newcommand{\zerodisplayskips}{%
  \setlength{\abovedisplayskip}{2pt}%
  \setlength{\belowdisplayskip}{2pt}%
  \setlength{\abovedisplayshortskip}{2pt}%
  \setlength{\belowdisplayshortskip}{2pt}}
\appto{\normalsize}{\zerodisplayskips}
\appto{\small}{\zerodisplayskips}
\appto{\footnotesize}{\zerodisplayskips}

\def\cvprPaperID{5640} 
\def\confName{CVPR}
\def\confYear{2023}

\begin{document}

\title{Unsupervised Gaze-aware Contrastive Learning \\with Subject-specific Condition}
\newcommand{\methodname}{ConGaze\xspace}
\author{Lingyu Du, Xucong Zhang, Guohao Lan\\
Delft University of Technology\\
{\tt\small \{Lingyu.Du, Xucong.Zhang, G.Lan\}@tudelft.nl}
}
\maketitle

\begin{abstract}
Appearance-based gaze estimation has shown great promise in many applications by using a single general-purpose camera as the input device. However, its success is highly depending on the availability of large-scale well-annotated gaze datasets, which are sparse and expensive to collect. To alleviate this challenge we propose \methodname, a contrastive learning-based framework that leverages unlabeled facial images to learn generic gaze-aware representations across subjects in an unsupervised way. Specifically, we introduce the gaze-specific data augmentation to preserve the gaze-semantic features and maintain the gaze consistency, which are proven to be crucial for effective contrastive gaze representation learning. Moreover, we devise a novel subject-conditional projection module that encourages a share feature extractor to learn gaze-aware and generic representations. Our experiments on three public gaze estimation datasets show that \methodname outperforms existing unsupervised learning solutions by 6.7\% to 22.5\%; and achieves 15.1\% to 24.6\% improvement over its supervised learning-based counterpart in cross-dataset evaluations.



\end{abstract}

\section{Introduction}
Gaze estimation is a technique of measuring where someone is looking at. It plays a key role in a wide range of applications including human-computer interaction \cite{majaranta2014eye}, the early detection of Autism spectrum disorder~\cite{shishido2019application,guillon2014visual}, mental workload estimation~\cite{yamada2018detecting,pfleging2016model}, and foveated rendering~\cite{patney2016towards,kim2019foveated} in virtual and augmented reality.
The computer vision community has made significant efforts to enable automatic and highly accurate gaze estimation systems~\cite{cheng2021appearance,ghosh2021automatic}.
Recently, appearance-based gaze estimation has regained 
its popularity after been proposed for two decades~\cite{cheng2021appearance}.
Compared to its model-based counterpart~\cite{hansen2009eye},
appearance-based gaze estimation is unobtrusive and can leverage the general-purpose cameras, such as those embedded in smartphones~\cite{krafka2016eye} and public displays~\cite{sugano2016aggregaze,zhang2018training}, to estimate the gaze directly from the captured facial images. Despite its great promise, the recent success of appearance-based gaze estimation is built on data-driven deep learning models and the availability of high-quality gaze datasets. However, collecting large-scale eye-tracking data with accurately annotated gaze labels is a labor-intensive and time-consuming process, which requires highly sophisticated setups and subject recruitment~\cite{Zhang2020ETHXGaze,zhang2015appearance,krafka2016eye,kellnhofer2019gaze360}. Although synthetic models~\cite{sugano2014learning,wood2015rendering,wood2016learning} have been proposed as data-efficient alternatives, there is a domain gap between the real and synthesized eye images~\cite{wang2022contrastive}.

\begin{figure}[]
	\centering
	\includegraphics[width=7.7cm]{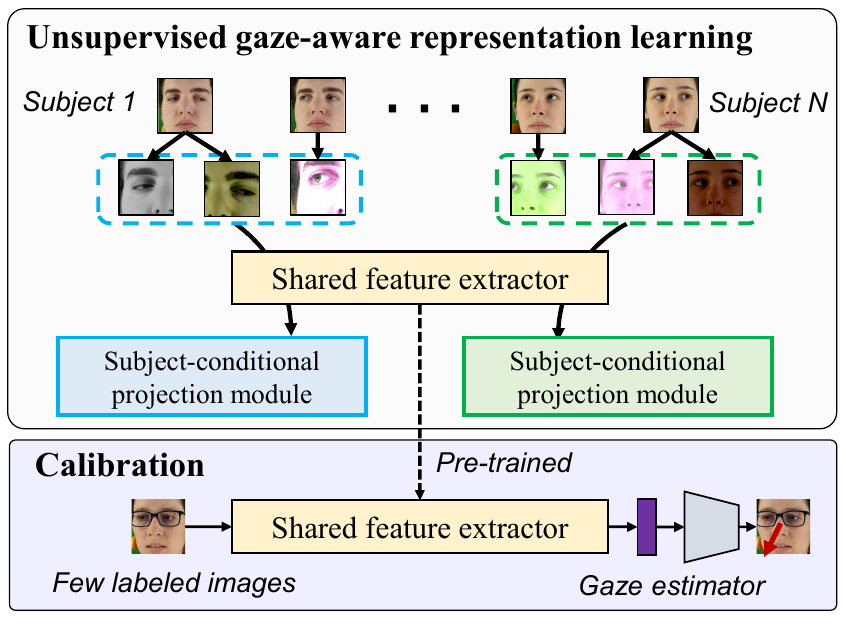}
	\caption{Illustration of \methodname, a contrastive learning-based framework for unsupervised gaze-aware representation learning. It consists of gaze-specific data augmentation and the subject-conditional projection module to encourage the learning of gaze-aware representations that are invariant to subject diversity. After unsupervised training, we only need a few labeled images in the calibration stage to fine-tune a simply neural networks for gaze estimation. 
	}
	\label{fig:teasing_figure}
	\vspace{-0.15in}
\end{figure}

In fact, lacking large-scale labeled dataset is a common challenge for many deep learning-based computer vision tasks \cite{wu2018unsupervised}. To alleviate this burden, unsupervised learning has received increasing attention from the community and has shown its potential in capturing effective representations from unlabeled data~\cite{hjelm2018learning,gidaris2018unsupervised,oord2018representation}. Among numerous unsupervised learning methods, contrastive learning~\cite{chen2020simple,chen2020improved,He_2020_CVPR,wu2018unsupervised} is dominating recent advances, and has achieved great performances in various downstream tasks including image classification~\cite{chen2020simple,dosovitskiy2014discriminative} and object detection~\cite{xie2021detco}. 

However, existing contrastive learning methods cannot directly accommodate to gaze estimation. Their common objective is to learn a feature extractor that ensures the representations of visually similar images, i.e., images containing the same instance or instances of the same category~\cite{wang2016contextual}, are close to each other in the embedding space~\cite{hadsell2006dimensionality}, while the representations of visually distinct images are apart from each other~\cite{ye2019unsupervised}. As a result, they tend to learn appearance-related semantic features that are beneficial to the mainstream classification task~\cite{wang2022contrastive}. Unfortunately, images with the same gaze label can be visually distinct for gaze estimation, i.e., facial images of different subjects sharing the same gaze. This special characteristic makes existing appearance-focused contrastive learning methods ill-suited for gaze estimation.

We tackle the challenge by proposing \methodname, a contrastive learning-based framework that leverages unlabeled full-face images to learn generic gaze-aware representations. Specifically, as shown in Figure~\ref{fig:teasing_figure}, 
we introduce {the gaze-specific data augmentation} to construct {gaze-consistent} and {gaze-contrastive pairs} for effective contrastive learning. The gaze-consistent pair contains two views augmented from the same full-face image, for which the gaze-related semantic features are {consistent}; while 
the gaze-contrastive pair includes two views augmented from two different images of the same subject, containing contrastive gaze-related features. Distinct from conventional data augmentations~\cite{chen2020simple}, our design preserves the gaze-semantic features and gaze consistency that are beneficial for contrastive gaze representation learning. 

Taking the image pairs from gaze-specific data augmentation as input, we devise a novel {subject-conditional projection module} that encourages a shared feature extractor to learn gaze-aware and generic features across different subjects. Different from conventional contrastive learning, which projects the learned representations to a single embedding space, we construct multiple subject-specific embedding spaces to perform contrastive learning. By doing so, we can accommodate identity and appearance-related features in each individual subject-specific embedding space, while encouraging the shared feature extractor to capture subject-invariant and gaze-related features in the shared common space.

Finally, we evaluate \methodname on three public datasets and compare it with state-of-the-art unsupervised methods~\cite{yu2020unsupervised,sun2021cross}. The results show that \methodname outperforms existing solutions in all examined scenarios, i.e., 22.5\% improvement over the method of Yu et al.~\cite{yu2020unsupervised}, and 6.7\% to 14.1\% improvement over the method of Sun et al.~\cite{sun2021cross} in different settings. Moreover, in cross-dataset scenarios, the performance of \methodname is 15.1\% to 24.6\% better than its supervised learning-based counterpart, which demonstrates \methodname can learn more robust and general representations that are transferable among datasets (subjects).

Our key contributions are threefold:
\begin{itemize} [leftmargin=*, wide, labelwidth=!, labelindent=0pt]

\item We propose \methodname, a novel contrastive learning-based framework to learn the gaze-aware representations that are generic across subjects. 

\item \methodname consists of the gaze-specific data augmentation and the subject-conditional projection module to endow contrastive learning 
for downstream gaze estimation task.

\item We conduct extensive experiments on three public datasets to demonstrate the superiority of \methodname to state-of-the-art methods.

\end{itemize} 

We will make the implementation of our framework available upon publication.




\section{Related Work}
\label{sec:RelatedWork}

\begin{figure*}[]
	\centering
	\includegraphics[width=16cm]{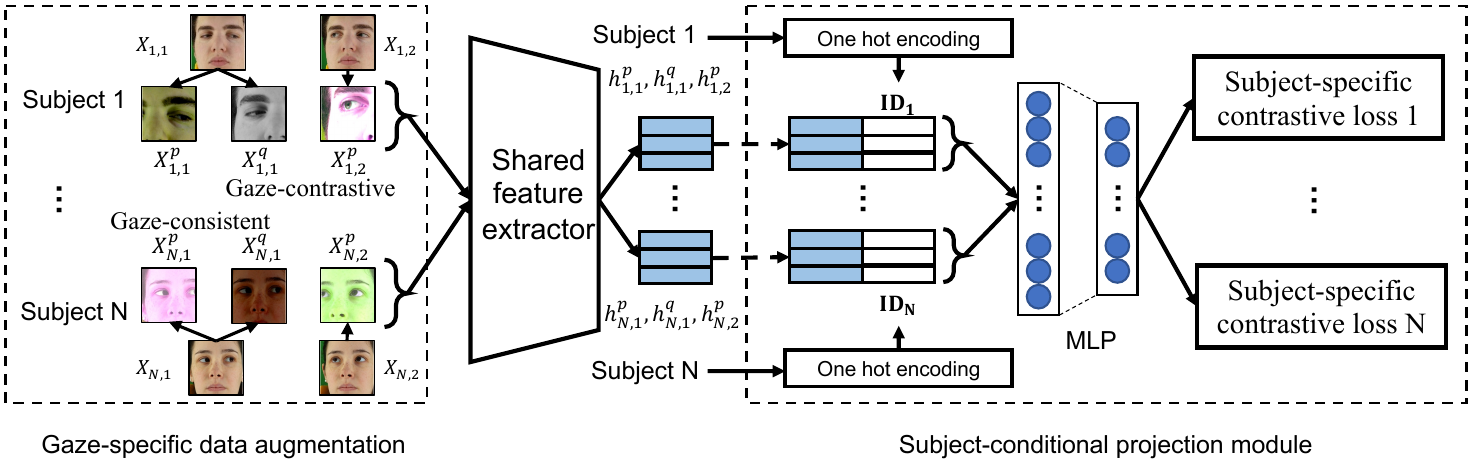}
	\caption{{The overview of \methodname}. 
	Taking unlabeled full-face images as inputs, we apply gaze-specific data augmentation to generate gaze-consistent pairs (the image pairs augmented from the same image), and gaze-contrastive pairs (the image pairs augmented from different images of the same subject). The augmented images are fed into the shared feature extractor to learn generic gaze-aware representations. To accommodate the subject-specific attributes, we design the subject-conditional projection module to map the general representations to multiple subject-specific embedding spaces. We apply the subject-specific contrastive loss on each subject-specific embedding space to encourage the learning of generic gaze-aware features. 
	}
	\label{fig:Framework}
	\vspace{-0.15in}
\end{figure*}


\textbf{Gaze Estimation}. Gaze estimation methods can be broadly classified into model-based 
and appearance-based methods. Model-based methods \cite{hansen2009eye} construct geometric models for gaze estimation by leveraging eye images captured by dedicated eye-tracking cameras \cite{nakazawa2012point,guestrin2006general,hansen2009eye}. 
In contrast, appearance-based methods directly retrieve gaze direction from facial images or eye images captured by conventional cameras. 
Earlier works in appearance-based gaze estimation only take eye images as the inputs~\cite{he2015omeg,huang2017tabletgaze,zhang2015appearance}. 
Recently, Krafka et al.~\cite{krafka2016eye} show that information in facial region can improve gaze estimation accuracy. Zhang et al.~\cite{zhang17_cvprw} propose an appearance-based gaze estimation method that takes full-face images for gaze estimation. 

As the same as many other computer vision tasks, appearance-based gaze estimation~\cite{Park_2019_ICCV,Yu_2019_CVPR,Zhu_2017_ICCV} benefits greatly from data-driven deep learning. Their performance is highly dependent on the availability of a large-scale dataset with gaze labels for supervised learning. Unfortunately, collecting large-scale dataset with accurate gaze labels is complicated and expensive \cite{yu2020unsupervised,Zhang2020ETHXGaze}, which becomes the fundamental barrier to the wide adoption of gaze estimation. Our work aims to mitigate this challenge by learning gaze-aware representations without access to gaze labels.


\textbf{Contrastive Learning.}
Contrastive learning~\cite{hao2021invariant,tian2020contrastive,bachman2019learning, pmlr-v119-henaff20a} has achieved great performance for 
many computer vision tasks, including image classification \cite{dosovitskiy2014discriminative} and object detection \cite{xie2021detco}. 
However, conventional contrastive learning methods focus on learning appearance-related representations, leading to poor performance on gaze estimation \cite{sun2021cross}. 
To solve this problem, we propose a suite of techniques that optimize contrastive learning for gaze estimation, which learns gaze-aware representations rather than appearance-related representations.

\textbf{Unsupervised Gaze Representation Learning.}
Yu et al.~\cite{yu2020unsupervised} propose an unsupervised gaze representation learning that jointly trains a gaze representation network and a gaze redirection network, such that the difference between the representations extracted from two eye images of the same subject can guide the gaze redirection network to map from one eye image to another eye image. 
More recently, Sun et al.~\cite{sun2021cross} introduce cross-encoder to disentangle gaze-aware representations and eye-identity representations by applying a latent-code-swapping mechanism on gaze-similar and eye-consistent image pairs. However, these two methods only take eye images as inputs and discard the facial region that contains useful information for gaze estimation \cite{zhang17_cvprw}. In contrast, our method can benefit from full-face images to achieve better performance.
%

\section{Method}
We propose \methodname, a contrastive learning-based framework that leverages unlabeled facial images to learn generic gaze-aware representations across subjects. 
Figure~\ref{fig:Framework} shows the overview of \methodname. First, we design the \textit{gaze-specific data augmentation module}, which takes unlabeled full-face images of multiple subjects as inputs, to create {gaze-consistent pairs} 
and {gaze-contrastive pairs}. 
Next, we feed the constructed image pairs to the \textit{shared feature extractor} to learn generic representations that are invariant to subject difference. Meanwhile, to encode subject-specific features, we design the \textit{subject-conditional projection} to further map the generic representations generated by the shared feature extractor to different subject-specific embedding spaces. 
Finally, to encourage the shared feature extractor to learn gaze-aware representations, we apply the \textit{subject-specific contrastive loss} in each subject-specific embedding space when training the end-to-end framework. We present the details of each component below.


\subsection{Gaze-specific Data Augmentation}
\label{GazeDataAugmentation}

\noindent\textbf{Rationale behind the design.} Data augmentation transforms a single image into two correlated views. The views transformed from the same image are considered as a positive pair, while views augmented from different images are considered as a negative pair~\cite{He_2020_CVPR}. In contrastive learning, representations are learnt by jointly training a feature extractor and a nonlinear projection head to solve the contrastive prediction task~\cite{chen2020simple,wu2018unsupervised}, which projects the representations of the positive pairs closer on an embedding space; while separates the representations of the negative pairs apart. By doing so, the feature extractor can capture representations that are consistent among the positive pairs, and are invariant to the augmentations that have been applied~\cite{Chuang_2022_CVPR,xiao2020should}. In the context of gaze estimation, we argue that the consistent representations among positive pairs should relate to and maintain the same gaze-semantic features. However, conventional data augmentation operators, such as cropping and cutout \cite{chen2020simple}, can remove the eyes from the image, while the others, such as rotation and flipping~\cite{komodakis2018unsupervised}, can change the gaze directions. As a result, 
they will generate noisy image views that fail to maintain the gaze-semantic consistency 
and lead to deteriorate effects. 
\vspace{0.1in}


\noindent \textbf{Detailed design.} To this end, we design {the gaze-specific data augmentation module} to create \textit{gaze-consistent pairs} and \textit{gaze-contrastive pairs} for effective contrastive gaze-aware representation learning. We define the gaze-consistent pair as two different views augmented from the same full-face image, for which the gaze-related semantic features are {consistent} and {invariant to the data augmentations} that have been applied. We define the gaze-contrastive pair as two views augmented from two different images of the same subject, containing contrastive gaze-related features. 

\begin{figure}[]
	\centering
	\includegraphics[width=8.2cm]{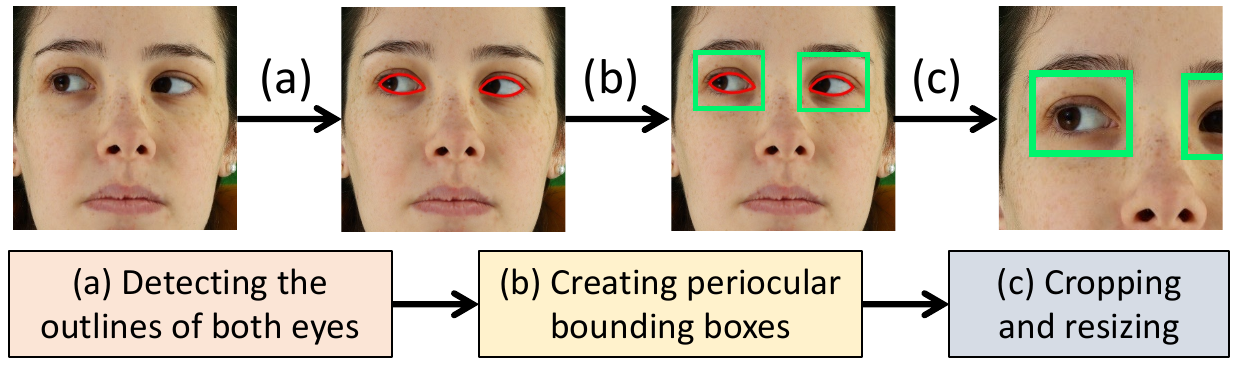}
	\caption{{The pipeline of gaze-cropping}: (a) detect the outlines of eyes; (b) create periocular bounding boxes; (c) randomly crop the image such that at least one periocular area is preserved in augmented view. 
	The augmented image maintains the gaze-semantic features and ensures the gaze consistency, i.e., subject in the original and augmented views has the same gaze direction. 
	}
	\label{fig:GazeCropping}
	\vspace{-0.15in}
\end{figure}


To construct the image pairs, we first propose a gaze-specific data augmentation operator, named gaze-cropping, that is able to maintain the gaze-related semantic features in the augmented positive image pairs. The pipeline of gaze-cropping is shown in Figure~\ref{fig:GazeCropping}. We first apply the facial landmark detection \cite{google} to obtain the outlines of both eyes. Then, based on detected outlines 
we create two bounding boxes that contain the two periocular areas, which has been proved to be valuable for gaze estimation \cite{zhang17_cvprw}. Next, we randomly crop the image based on the positions of the bounding boxes such that at least one periocular area is preserved in the cropped images. Finally, the cropped image is resized to the size of the original image. 

Inspired by the fact that the composition of color distortion and cropping-based operators can improve the effectiveness of contrastive learning~\cite{chen2020simple}, we adopt color distortion as the second data augmentation operator. The final design of the gaze-specific data augmentation is the composition of gaze-cropping and color distortion, for which we apply the two data augmentation operators sequentially on the image to generate gaze-consistent pairs. 

Formally, we denote the gaze-specific data augmentation as a random variable $\mathcal{T}$, and denote the $j$th image of subject $i$ as $X_{i,j}$. Each time the data augmentation is performed by randomly sampling two augmentation operators from the random variable $\mathcal{T}$, i.e., $\{p,q\}\sim \mathcal{T}$. Then, a gaze-consistent pair $(X_{i,j}^p, X_{i,j}^q)$ is generated by applying the sampled operators $p$ and $q$ to $X_{i,j}$. Similarly, a gaze-contrastive pair $(X_{i,j}^p, X_{i,k}^q)$ is constructed by applying $p$ and $q$ to different images $X_{i,j}$ and $X_{i,k}$ from subject $i$. The left part of Figure~\ref{fig:Framework} shows some examples of the constructed gaze-consistent and gaze-contrastive pairs from two subjects. 






\subsection{Subject-conditional Projection Module}



In conventional contrastive learning~\cite{chen2020simple}, a feature extractor $\textbf{F}(\cdot)$ is used to map the augmented image $X$ into a 
general representation space $\mathbb{GP}$ and extracts representations $\mathbf{h}=\textbf{F}(X)\in \mathbb{GP}$. Then, a common projection head $\textbf{P}(\cdot)$ further maps the extracted representations $\mathbf{h}$ into a 
embedding space $\mathbb{SP}$ to obtain embeddings $\mathbf{z}=\textbf{P}(\mathbf{h})\in \mathbb{SP}$. In general, $\textbf{P}(\cdot)$ is implemented as a simple multi-layer perceptron (MLP) with one hidden layer~\cite{chen2020simple}. The contrastive loss is applied in the embedding space $\mathbb{SP}$, and a commonly used contrastive loss is the InfoNCE~\cite{oord2018representation}. 

The two neural networks $\textbf{F}(\cdot)$ and $\textbf{P}(\cdot)$ are jointly trained by solving the contrastive prediction task, such that a well-trained $\textbf{F}(\cdot)$ can extract $\mathbf{h}$ to distinguish images among each other~\cite{wu2018unsupervised}. However, as the contrastive loss is defined in the common embedding space $\mathbb{SP}$ and the mini-batch contains full-face images sampled from different subjects, these two factors will have negative impacts on the downstream gaze estimation task. The reason is fairly intuitive as the differences in the subject appearance will encourage $\textbf{F}(\cdot)$ to learn appearance-related or identity-related semantic features that can be easily learnt and used to solve the contrastive prediction task (differentiating images of different subjects) defined in $\mathbb{SP}$. Thus, conventional design fails to learn gaze-aware features that are beneficial for gaze estimation. 

We resolve this problem by designing and optimizing multiple subject-specific contrastive losses using a subject-conditional projection module with a shared feature extractor. We introduce our design below.

 \begin{figure}[]
	\centering
	\includegraphics[width=8.cm]{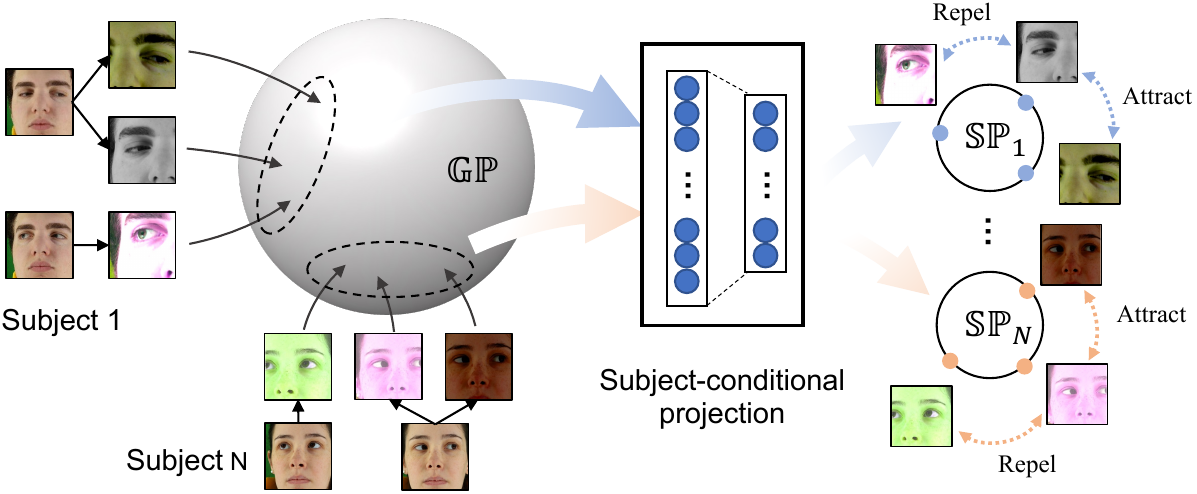}
	\caption{{Illustration of the subject-conditional projection}. The subject-conditional projection maps the representations learned by the shared feature extractor from the general representation space $\mathbb{GP}$ to $N$ subject-specific embedding spaces ${\mathbb{SP}_i}$, where $i\in [1, N]$ and $N$ is the number of subjects. In each $\mathbb{SP}_i$, the views in the gaze-consistent pair attract each other, whereas the images in the gaze-contrastive pair repel each other. 
	}
	\label{fig:latentSpace}
	\vspace{-0.15in}
\end{figure}


\subsubsection{Subject-conditional Projection} 

As shown in Figure \ref{fig:latentSpace}, instead of mapping all images into the common embedding space $\mathbb{SP}$ that is invariant to different subjects, we propose the subject-conditional projection $\textbf{S}(\cdot)$ to construct $N$ separate embedding spaces, where $N$ is the number of subjects appeared in the training dataset. Each embedding space $\mathbb{SP}_i$ is specialized to a single subject $i$, and is able to {accommodate subject-specific embeddings}, i.e., appearance and identity-related features. 

When solving the contrastive prediction task in each embedding space $\mathbb{SP}_i$, $\textbf{F}(\cdot)$ needs to maximize the similarity of the embeddings of the gaze-consistent pairs, while minimize that of the gaze-contrastive pairs augmented from the same subject. By doing so, it encourages $\textbf{F}(\cdot)$ to learn gaze-aware representations, as the appearance and identity-related semantic features are not useful for solving the contrastive prediction task, i.e., differentiating images of the same subject containing different gaze features. The detailed design of the subject-conditional projection module is shown in the right part of Figure \ref{fig:Framework}. Specifically, for each augmented image $X_{i,j}^t, t\in \{p,q\}$, we first apply one hot encoding to encode the identity information. The encoded identity information is denoted by $\textbf{ID}_i$. Then, we concatenate $\textbf{ID}_i$ with the representations $\mathbf{h}_{i,j}^t=\textbf{F}(X_{i,j}^t)\in \mathbb{GP}$ that are extracted by 
$\textbf{F}(\cdot)$. Finally, 
$\{\textbf{ID}_i, h_{i,j}^t\}$ is fed into a MLP to obtain the subject-specific embeddings $\mathbf{z}_{i,j}^t\in \mathbb{SP}_i$. The whole process of subject-conditional projection is denoted by $\mathbf{z}_{i,j}^t = \textbf{S}(\mathbf{h}_{i,j}^t)$.

\subsubsection{Subject-specific Contrastive Loss} 

We define the subject-specific contrastive loss on each 
embedding space $\mathbb{SP}_i$. Formally, for subject $i$, we define the subject-specific contrastive loss for a gaze-consistent pair $(X_{i,j}^p, X_{i,j}^q)$ as:
\begin{align}
\mathcal{L}_{i, j} = -\log{\frac{\exp \Big(\text{sim}(\mathbf{z}_{i,j}^p,\mathbf{z}_{i,j}^q)/\tau\Big)}{\sum_{k=1}^{K_i}{\exp\Big(\text{sim}(\mathbf{z}_{i,j}^p,\mathbf{z}_{i,k}^q)/\tau\Big)}}},
\label{contrastiveLoss}	
\end{align}
where $\mathbf{z}_{i,j}^p$ and $\mathbf{z}_{i,j}^q$ are the subject-specific embeddings of the gaze-consistent pair; $\mathbf{z}_{i,j}^p$ and $\mathbf{z}_{i,k}^q (k \neq j)$ are the subject-specific embeddings of a gaze-contrastive pair; $\text{sim}(u,v)=u^\top v/\|u\|\|v\|$ is the cosine similarity of two feature vectors $u$ and $v$; $K_i$ is the number of images for subject $i$ in the minibatch; and $\tau$ is a temperature parameter. The final subject-specific contrastive loss for subject $i$ is computed over all gaze-consistent pairs constructed from the mini-batch. 


Since the subject-specific contrastive loss is defined in each individual embedding space $\mathbb{SP}_i$, it allows $\mathbb{SP}_i$ to accommodate the subject-specific features. Meanwhile, optimizing multiple contrastive losses can encourage the shared feature extractor to learn representations in the shared common space $\mathbb{GP}$ that are invariant to subject diversity and augmentation.

\section{Evaluation}


We first compare the performance of \methodname with state-of-the-art methods~\cite{yu2020unsupervised,sun2021cross}. We then conduct ablation study to evaluate the effectiveness of the proposed gaze-specific data augmentation and the subject-conditional projection module in learning generic gaze-aware representations. 

\subsection{Datasets}

We consider three public datasets 
in the evaluation: 


\noindent \textbf{ColumbiaGaze (C)}~\cite{CAVE_0324} is collected from 56 subjects. 
For each subject, 105 facial images are collected from seven horizontal and three vertical gaze directions. 
We crop the face patches from the original images and resize them to $224\times224$. 
\vspace{0.1in}

\noindent \textbf{MPIIFaceGaze (M)}~\cite{zhang17_cvprw} is collected from 15 subjects in their daily life under different illumination conditions, head poses, and backgrounds. Each subject has 3,000 full-face images. We resize the original images to the resolution of $224\times224$. 
\vspace{0.1in}


\noindent \textbf{ETH-XGaze (E)}~\cite{Zhang2020ETHXGaze} is collected from 110 subjects with 18 head poses and different illumination conditions. It consists of a training set that includes 80 subjects, a testing set (TES) that contains 15 subjects for subject-specific evaluation, and a testing set (TE) that contains 15 subjects for within-dataset evaluation. Only the gaze labels for the training set are released. 
The image resolution is 224 $\times$ 224.


\subsection{Implementation}

We implement the proposed \methodname with Python and TensorFlow 2.0. We use ResNet-18 \cite{he2016deep} (without the dense layer) to implement the shared feature extractor. For the pre-training,  
we use Adam optimizer with a learning rate of 0.01. The batch size for dataset \textbf{C}, \textbf{M}, and \textbf{E} is 105, 128, and 128, respectively. 
In each training iteration, we randomly sample a batch of images from one subject and calculate the corresponding subject-specific contrastive loss to train the 
neural networks by performing gradient descent for one time. 


In the calibration stage, we use two fully-connected layers to implement the gaze estimator, which takes the representations generated by the shared feature extractor as inputs and outputs the predicted gaze direction. The neural network is fine-tuned for 150 epochs.

\subsection{{Comparison with State-of-the-art}}


Below, we compare \methodname with existing unsupervised gaze-aware representation learning methods \cite{yu2020unsupervised,sun2021cross} and the conventional contrastive learning method, i.e., SimCLR \cite{chen2020simple}. 
For a fair comparison, we use the same experimental settings as \cite{sun2021cross,yu2020unsupervised}, and consider both within-dataset and cross-dataset evaluations. The details of experimental setup are as follows:

\vspace{0.1in}
\noindent \textbf{Within-dataset.} We perform five-fold and leave-one-subject-out cross-validation on \textbf{C} and \textbf{M} datasets. In each fold, we first use the training set (without gaze annotations) to train the shared feature extractor. Then, we randomly select 100 image samples with gaze annotations from the training set for calibration. 
\vspace{0.1in}

\noindent\textbf{Cross-dataset.} In this setting, we pre-train the shared feature extractor on one dataset and evaluate it on the other. Specifically, we use the notation \textbf{X-Y} to indicate that the shared feature extractor is pre-trained on dataset \textbf{X} and evaluated on dataset \textbf{Y}. Similar to~\cite{sun2021cross}, we consider three cross-dataset evaluations, \textbf{C-M}, \textbf{E-C}, and \textbf{E-M}. For \textbf{E-C} and \textbf{E-M}, the shared feature extractor is pre-trained on the training set of \textbf{E}. We perform five-fold and leave-one-subject-out cross-validation on \textbf{C} and \textbf{M} respectively. 
In each fold, 100 samples with gaze labels are randomly selected from the training set for calibration. 
\vspace{0.1in}

For all settings, we repeat the evaluation ten times for each fold by randomly choosing different calibration samples. We report the averaged angular errors. 


\begin{table}[]
\caption{{The averaged angular error (in degree) for within-dataset and cross-dataset evaluations} of different methods. 
Note that for Sun et al.~\cite{sun2021cross} dataset \textbf{M} represents the MPIIGaze, which only contains the eye area patch. Moreover, Sun et al.~\cite{sun2021cross} first pre-train the feature extractor on ColumbiaGaze (C) in an unsupervised manner, and further train on MPIIGaze (M) for ten epochs. 
}
\label{PerformanceComparison}
\centering
\resizebox{3.2in}{!}{
\begin{tabular}{l|c|c|c|c|c}
\hline
\textbf{Method}       & \textbf{C} & \textbf{M} & \textbf{C-M} & \textbf{E-C} & \textbf{E-M} \\ \hline \hline
SimCLR        & 12.3                    & 11.5                     & 11.4            & 12.8         & 12.2         \\
Yu et al.   & 7.1                   & -                      & -            & -          & - \\
Sun et al. & 6.4                    & 7.5                    & 8.3          & 7.7  &9.0           \\
Early prototype        & \textbf{5.5}          & \textbf{7.0}           & \textbf{7.7}        & \textbf{7.2}    & \textbf{9.0}   \\ \hline
\end{tabular}
}
\vspace{-0.15in}
\end{table}


\textbf{Results.} Table~\ref{PerformanceComparison} compares the averaged angular error of different methods in both within-dataset and cross-dataset evaluations. We report the {angular}  
errors mentioned in Yu et al.~\cite{yu2020unsupervised} and Sun et al.~\cite{sun2021cross}. Note that for Sun et al.~\cite{sun2021cross}, dataset \textbf{M} is the MPIIGaze dataset~\cite{zhang19_pami}, as their method can only leverage the eye area patch for gaze estimation. Moreover, their result is obtained by first pre-training the feature extractor on the ColumbiaGaze dataset, and further trained on MPIIGaze for ten epochs. 

As shown in Table~\ref{PerformanceComparison}, in within-dataset evaluation \methodname achieves a 1.6$^\circ$ (22.5\%) improvement over the method of Yu et al. \cite{yu2020unsupervised} on dataset \textbf{C}. When compared with the method of Sun et al.~\cite{sun2021cross}, \methodname improves by 0.9$^\circ$ (14.1\%) and 0.5$^\circ$ (6.7\%) on \textbf{C} and \textbf{M} datasets, respectively. In cross-dataset evaluations, \methodname achieves 0.6$^\circ$ (7.2\%) and 0.5$^\circ$ (6.7\%) improvements on \textbf{C-M} and \textbf{E-C}, respectively, over the method of Sun et al.~\cite{sun2021cross}. Next, when compared with 
SimCLR, \methodname achieves significant 4.5$^\circ$-6.8$^\circ$ (3.91\%-55.3\%) and 3.2$^\circ$-5.6$^\circ$ (26.2\%-43.8\%) improvements in within-dataset and cross-dataset evaluations, respectively. Finally, the results also indicate that \methodname can learn more generic gaze-aware representations than existing solutions, as all the experiments are conducted in a cross-subject setting and \methodname achieves better performance when adapting to unseen subjects.

\textbf{Discussion.} Different from the methods of Yu et al.~\cite{yu2020unsupervised} and Sun et al.~\cite{sun2021cross} that take eye patches as the inputs, \methodname 
learns gaze-aware representations from full-face images, which allows it to capture more effective representations that can benefit the downstream gaze estimation task. SimCLR also leverages full-face images for learning, but performs significantly worse than \methodname in all examined cases. There are two reasons for this performance gap. First, \methodname leverages the proposed gaze-specific data augmentation to ensure the consistency of gaze-related features among gaze-consistent pairs, whereas the conventional data augmentation adopted by SimCLR cannot guarantee such consistency. Second, \methodname leverages the subject-conditional projection module, which encourages the shared feature extractor to focus on the learning of gaze-aware representations rather than identity or appearance-related representations.

\subsection{Ablation Study}
\label{AblationStudySec}

\begin{figure*}[]
	\centering
	\includegraphics[width=17cm]{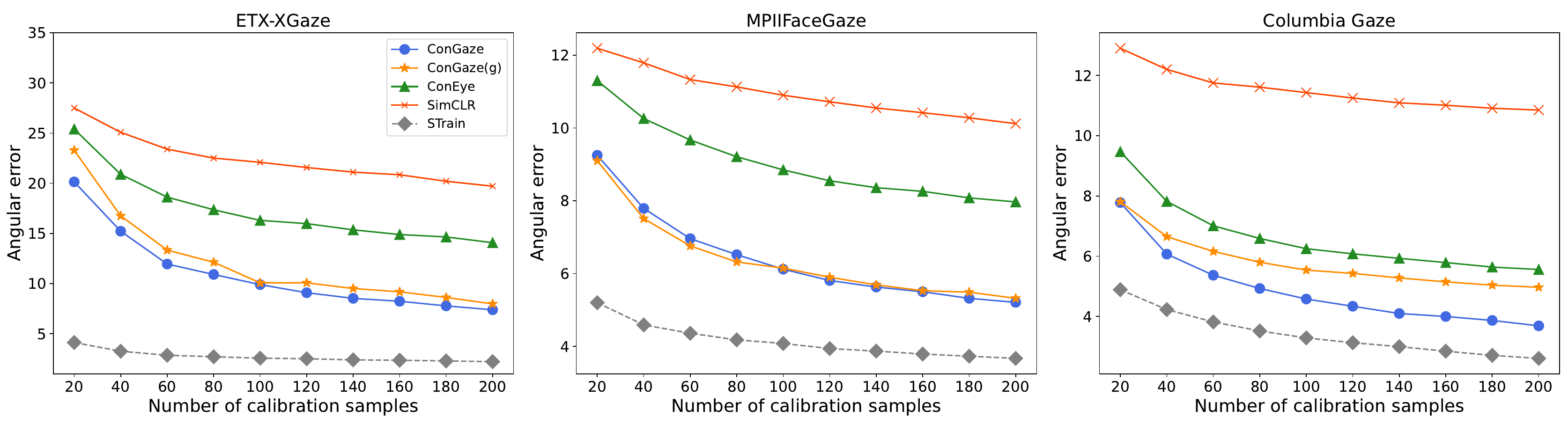}
	\caption{{Angular error when different amounts of samples are used during the calibration}.}
	\label{fig:varyingSize}
	\vspace{-0.15in}
\end{figure*}



Below, we study the impact of different design and training choices 
on the gaze estimation performance. Specifically, we consider the following methods: 

\begin{itemize} [leftmargin=*, wide, labelwidth=!, labelindent=0pt]

\item \textbf{RanNet}: 
both the shared feature extractor and the gaze estimator are randomly initialized. 
We train them with labeled samples in the calibration stage. 
\item \textbf{SimCLR}: the shared feature extractor is pre-trained by SimCLR, without applying the gaze-specific data augmentation and the subject-conditional projection module. In the calibration stage, we only fine-tune the parameters of the gaze estimator. 
\item \textbf{ConEye}: a variant of {SimCLR}, which adopts the proposed gaze-specific data augmentation.
\item \textbf{\methodname}: the shared feature extractor is pre-trained by the proposed method. In the calibration stage, we fine-tune the parameters of both the shared feature extractor and the gaze estimator.
\item \textbf{\methodname (g)}: different from \textbf{\methodname}, in the calibration stage, we only fine-tune the gaze estimator.
\item \textbf{STrain}: the shared feature extractor and the gaze estimator are pre-trained in supervised manner. 
In the calibration stage, we fine-tune both the shared feature extractor and the gaze estimator.
\end{itemize}

\textbf{Experiment setup.} We 
perform both within-dataset and cross-dataset evaluations. 

For {within-dataset evaluation}, we conduct five-fold cross-validation on datasets \textbf{C} and \textbf{M}. In the calibration stage, 100 images with gaze labels are randomly sampled from the testing set and the rest are used for evaluation. For dataset \textbf{E}, we use all the data in the training set to pre-train the shared feature extractor. Then, for each subject in the TES set, we use the first 100 images with gaze annotations for subject-specific gaze estimation. The final results are the averaged angular errors over 15 subjects in the TES set. For {cross-dataset evaluation}, we consider \textbf{E-C} and \textbf{E-M}, where the shared feature extractor is pre-trained on \textbf{E} and evaluated on \textbf{C} and \textbf{M} by five-fold cross-validation. The calibration images are sampled from the testing set. 
We repeat each experiment ten times 
and report the averaged angular errors. 


\textbf{Results.} Table~\ref{AblitionStudy} compares the averaged angular errors achieved by different methods in both within-dataset and cross-dataset evaluations. We make the following key observations. 

Among the four unsupervised learning methods, {SimCLR} has the highest angular error, which indicates that the conventional contrastive learning fails to learn gaze-aware representations. 
{ConEye} outperforms {SimCLR} in all cases with a $1.2^\circ$ to $5.8^\circ$ (10.9\% to 45.6\%) improvement, as the proposed gaze-specific data augmentation can maintain the consistency of gaze-related semantic features among gaze-consistent pairs and improve the quality of learned gaze-aware representations. After incorporating the subject-conditional projection module, {\methodname} achieves the lowest angular error among all unsupervised learning-based methods, and results in a $1.7^\circ$ to $7.0^\circ$ (25.5\% to 57.4\%) improvement over ConEye. Thus, we conclude that the subject-conditional projection module can encourage the shared feature extractor to learn gaze-aware representations. Finally, {RanNet} has the highest angular error on all the evaluation settings, as it does not involve any pre-training. {STrain} achieves the lowest angular error in most of the examined cases by taking advantage of labeled samples for supervised pre-training.

\begin{table}[]
\centering
\caption{{Ablation study} in both within-dataset and cross-dataset evaluations. The results are averaged angular errors when using 100 labeled samples for calibration.
}
\label{AblitionStudy}
\resizebox{2.8in}{!}{
\begin{tabular}{l|c|c|c|c|c}
\hline
\textbf{Method}     & \textbf{C} & \textbf{M} & \textbf{E} & \textbf{E-C} & \textbf{E-M} \\ \hline
\hline
RanNet     & 13.2 & 11.5 & 46.9 & -   & -   \\
SimCLR     & 11.4 & 10.9 & 22.1 & 13.4   & 11.0   \\
ConEye    & 6.2 & 8.8 & 16.3 & 12.2   & 9.8   \\
\methodname  & 4.5 & 6.1 & 9.9 & 5.2   & 7.3   \\
\methodname (g)    & 5.5 & 6.1 & 10.0 & 6.0   & 7.0   \\
STrain & 3.2 & 4.0 & 2.5 & 6.9   & 8.6   \\ \hline
\end{tabular}}
\vspace{-0.15in}
\end{table}


The fine-tuning method used in the calibration also influences the performance. As shown in Table~\ref{AblitionStudy}, \methodname has a similar performance with \methodname (g) on \textbf{M} and \textbf{E}, but improves the accuracy by 1.0$^\circ$ (18.2\%) and 0.8$^\circ$ (13.3\%) on \textbf{C} and \textbf{E-C}, respectively. Despite \methodname achieves a $0.3^\circ$ higher angular error on \textbf{E-M}, 
it benefits from fine-tuning both the shared feature extractor and the gaze estimator. 

For within-dataset evaluations, the performance gap between \methodname and 
STrain is $1.3^\circ$, $2.1^\circ$, $7.4^\circ$ on \textbf{C}, \textbf{M}, and \textbf{E}, respectively. {The gap is large on \textbf{E}, which we believe is due to the diverse head poses in ETH-XGaze.} Interestingly, for the cross-dataset evaluations, both \methodname and \methodname (g) perform significantly better than STrain. \methodname improves the estimation performance by 1.7$^\circ$ (24.6\%) and 1.3$^\circ$ (15.1\%) on \textbf{E-C} and \textbf{E-M}, respectively. The results indicate that the proposed framework can encourage the shared feature extractor to learn more generic representations compared with its supervised learning-based counterpart.


Finally, since unsupervised learning-based methods require calibration samples for fine-tuning, we examine the impact of different amounts of calibration samples on gaze estimation performance. The results are shown in Figure~\ref{fig:varyingSize}. 
The angular errors of all examined methods decrease as the number of calibration samples increase. The fully-supervised STrain performs better than all the other methods. 
The performance gap between \methodname and STrain reduces when more calibration samples are used.

\subsection{Visualizations}
\label{subsec:tsneVisualization}

\begin{figure}[]
	\centering
    \begin{subfigure}{0.32\linewidth}
        \includegraphics[width=1.in]{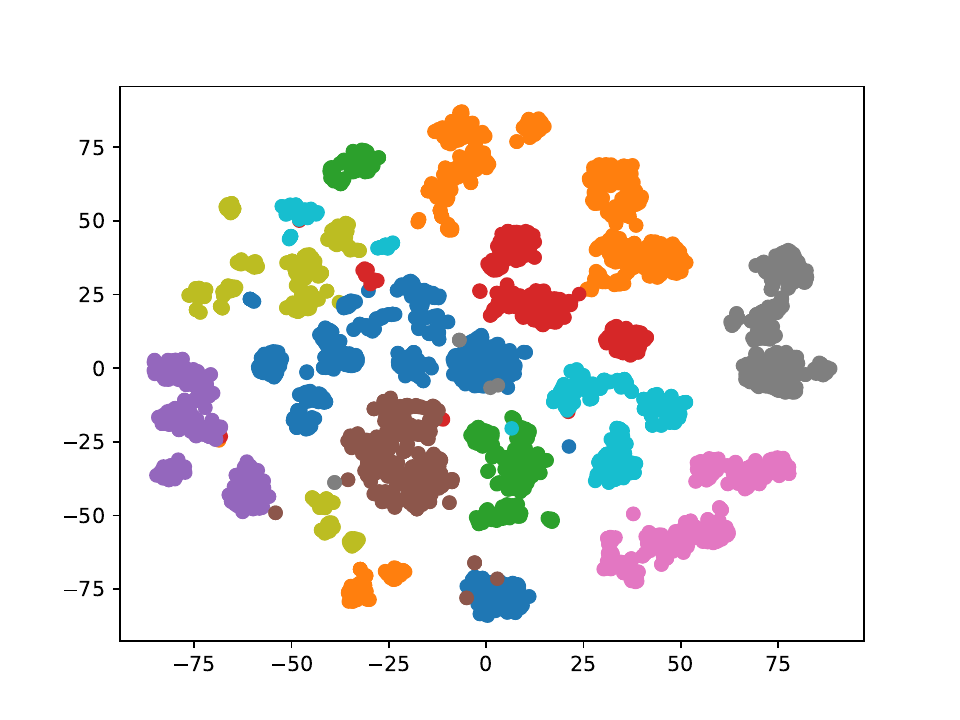}
        \caption{SimCLR on \textbf{M}}
    \end{subfigure}
    \begin{subfigure}{0.32\linewidth}
        \includegraphics[width=1.in]{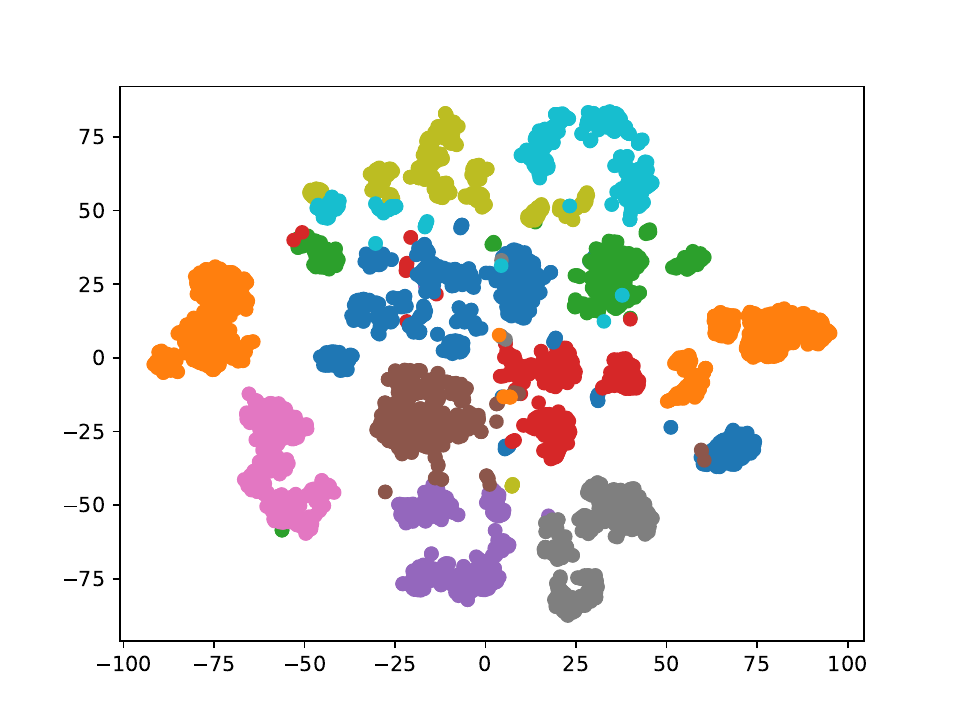}
        \caption{ConEye on \textbf{M}}
    \end{subfigure}
    \begin{subfigure}{0.32\linewidth}
        \includegraphics[width=1.in]{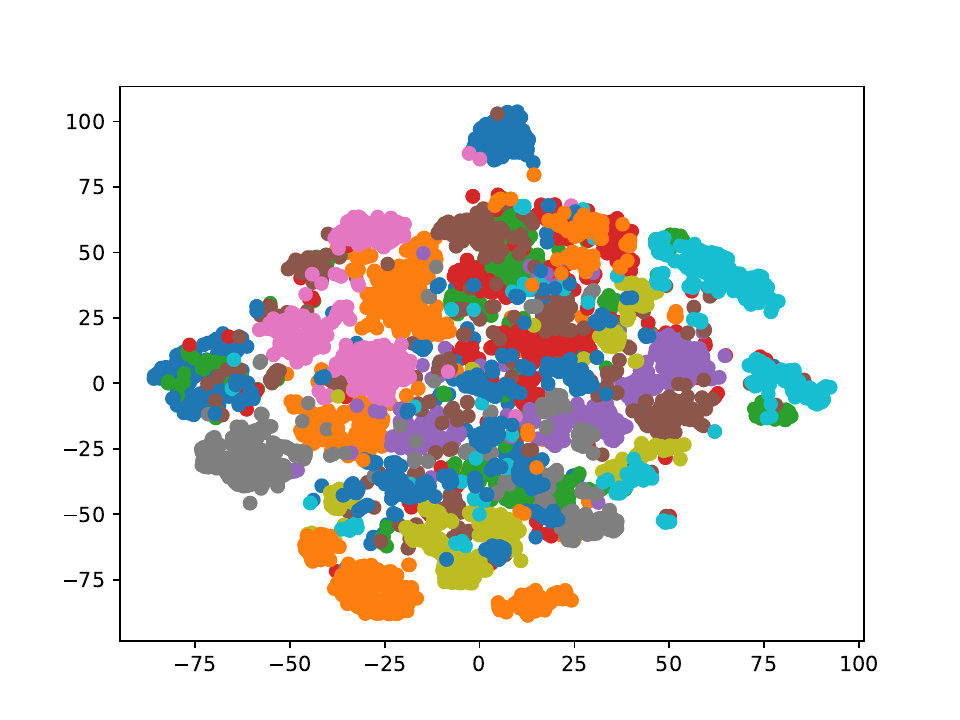}
        \caption{ConGaze on \textbf{M}} 
    \end{subfigure}
    \begin{subfigure}{0.32\linewidth}
        \includegraphics[width=1.15in]{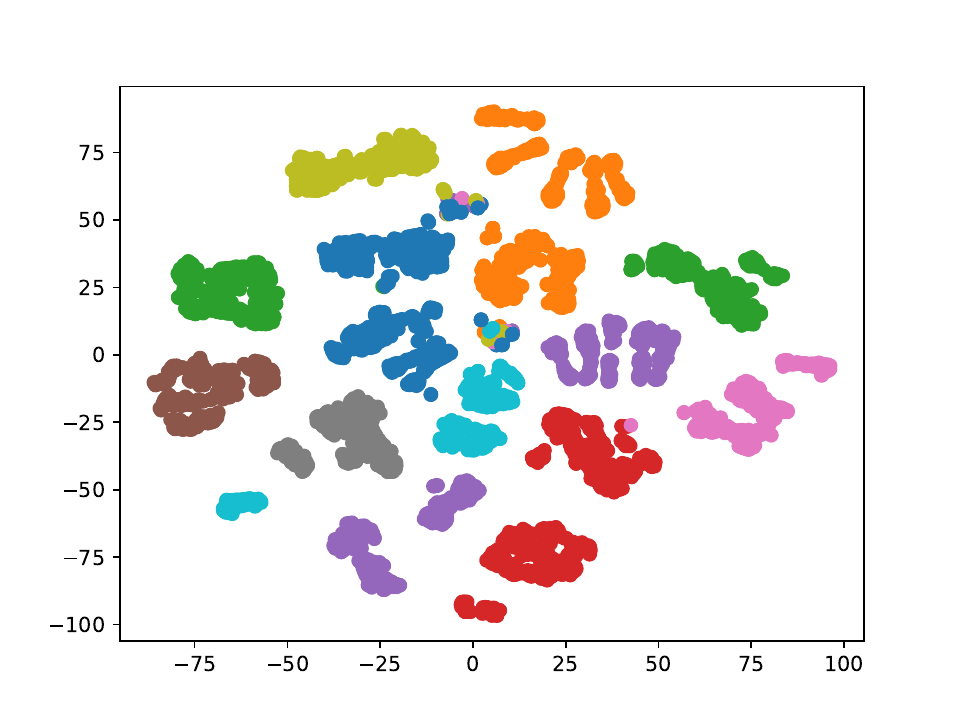}
        \caption{SimCLR on \textbf{E}}
    \end{subfigure}
    \begin{subfigure}{0.32\linewidth}
        \includegraphics[width=1.15in]{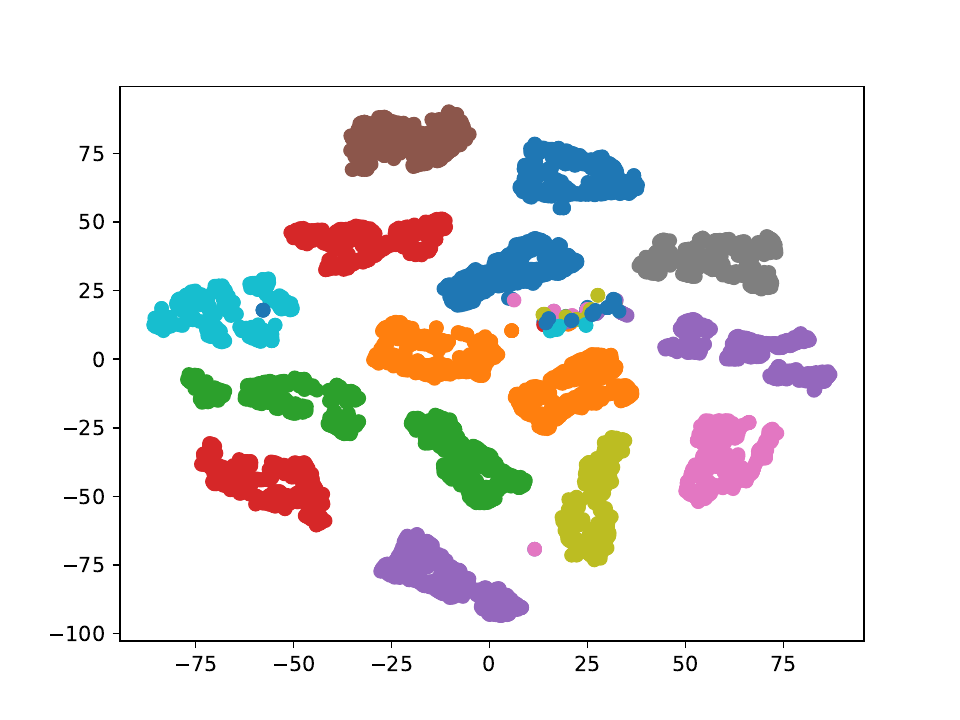}
        \caption{ConEye on \textbf{E}}
    \end{subfigure}
    \begin{subfigure}{0.32\linewidth}
        \includegraphics[width=1.15in]{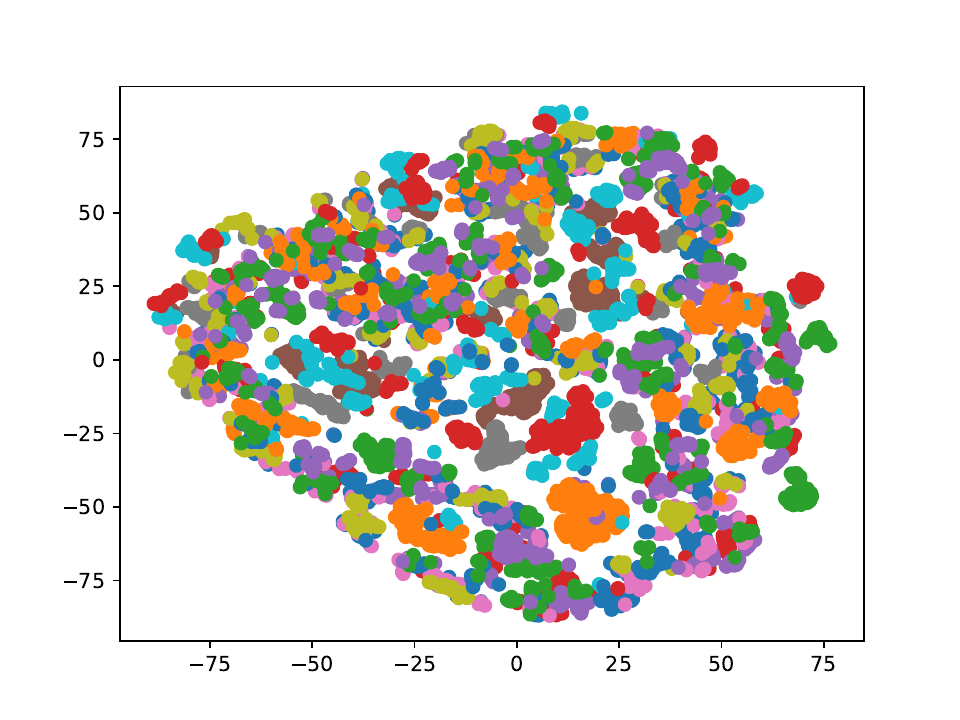}
        \caption{ConGaze on \textbf{E}} 
    \end{subfigure}
	\caption{{Visualization of the representations learned by SimCLR, ConEye, and \methodname using t-SNE} \cite{van2008visualizing} on datasets \textbf{M} and \textbf{E}. Different colors correspond to different subjects, and each point corresponds to one image. 
	} 
	\label{fig:tsneCompare}
	\vspace{-0.1in}
\end{figure}

\noindent\textbf{Visualization of representations.} To understand the effectiveness of the proposed {shared feature extractor} and subject-conditional projection module, we visualize the representations learned by SimCLR, ConEye, and \methodname in the general representation space $\mathbb{GP}$. We use t-SNE~\cite{van2008visualizing} to calculate the 2-dimensional features on both MPIIFaceGaze (\textbf{M}) and ETH-XGaze (\textbf{E}). 
The results are shown in Figure~\ref{fig:tsneCompare}. Different colors correspond to different subjects, and each data point corresponds to representation of one image. We observe that the representations learned by SimCLR and ConEye are well-clustered by subject identity, whereas the representations learned by \methodname are dispersed and highly overlapped. This demonstrates the effectiveness of \methodname in learning general gaze-aware representations that are invariant to subject identity. It also explains why \methodname achieves better performance in the cross-dataset evaluations (shown in Table~\ref{AblitionStudy}), as the learned generic representations are more robust to subject diversity and can be easily adapted to unseen subjects.
 
\begin{figure}[]
	\centering
	\includegraphics[width=7.cm]{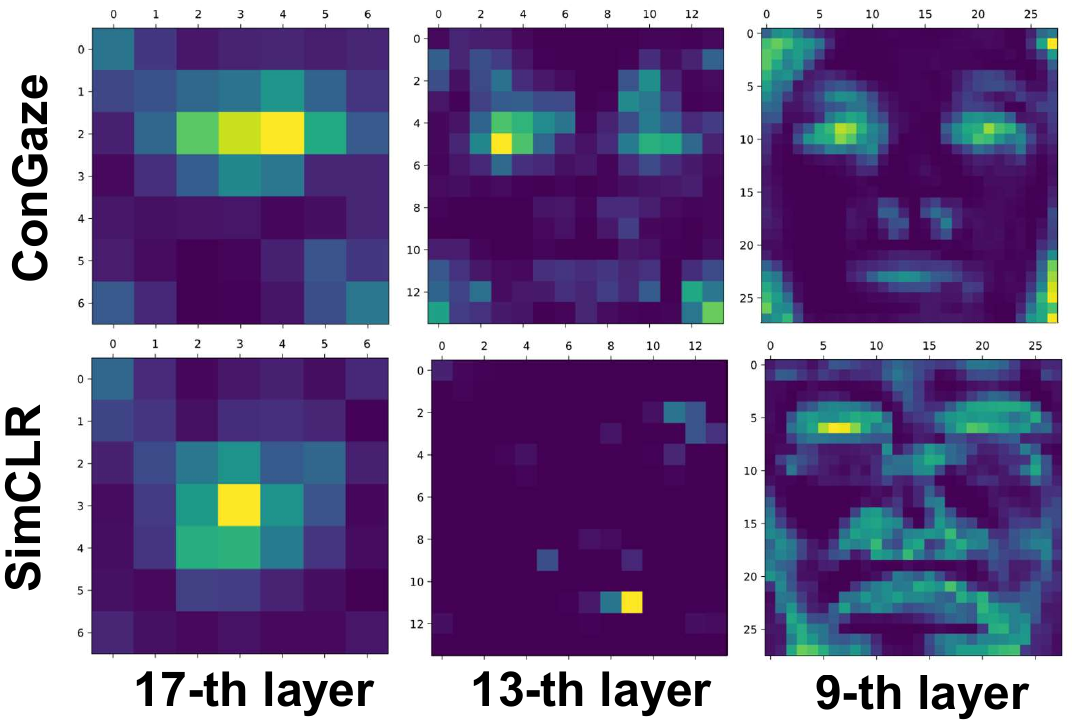}
	\caption{Visualization of the attention area obtained at different layers of the shared feature extractor. The two rows correspond to the shared feature extractor pre-trained by SimCLR and ConGaze, respectively.
	}
	\label{fig:heatmap}
	\vspace{-0.2in}
\end{figure}

\vspace{0.1in}
\noindent \textbf{Visualization of attention area.} To further demonstrate the effectiveness of \methodname in learning gaze-aware features, we compare the attention maps generated by the shared feature extractor that is pre-trained by \methodname and SimCLR, respectively. Specifically, we visualize the attention area of the $9$th, $13$th, and $17$th convolutional layers of the shared feature extractor using Grad-cam~\cite{Selvaraju_2017_ICCV} in Figure~\ref{fig:heatmap}. 
We can notice that SimCLR focuses on appearance-related features that are evenly spread over the facial images, whereas \methodname pays more attention to the periocular, nose, and mouth regions in the face that have been proven to be effective for gaze estimation~\cite{zhang17_cvprw}. This observation is clear in all the three layers we examined in Figure~\ref{fig:heatmap}. 

\section{Conclusion}
We propose \methodname, an unsupervised learning framework for the gaze estimation task leveraging contrastive learning.
To adapt the conventional contrastive learning for the gaze estimation task, we largely improve the original data augmentation in contrastive learning to the gaze-aware augmentation and propose the subject-conditional projection module to guide the learning of generic gaze-aware feature across subjects. Our extensive experiments show that \methodname achieves the better performance than previous state-of-the-art methods on multiple datasets. We believe that \methodname can reduce the burden of relying on large-amount labeled data for gaze estimation.


\balance

{\small
\bibliographystyle{ieee_fullname}
\bibliography{08_references}
}

\end{document}